\pdfoutput=1

\documentclass[11pt]{article}

\usepackage[]{EMNLP2023}

\usepackage{times}
\usepackage{latexsym}

\usepackage[T1]{fontenc}

\usepackage[utf8]{inputenc}

\usepackage{microtype}

\usepackage{inconsolata}

\usepackage{graphicx}
\usepackage{listings}
\usepackage{amsthm}
\usepackage{csquotes}
\usepackage{verbatim}
\usepackage{subcaption}
\usepackage{multirow,multicol}
\usepackage{amsmath,amsfonts,amssymb,amsthm}
\usepackage{booktabs}
\theoremstyle{definition}

\usepackage[linesnumbered,ruled,vlined]{algorithm2e}
\SetKwInput{KwInput}{Input}
\SetKwInput{KwOutput}{Output}
\usepackage{graphicx}
\usepackage{caption}
\usepackage{subcaption}
\usepackage{url}

\definecolor{mygreen}{rgb}{0.3, 0.78, 0.65}

\title{Relabeling Minimal Training Subset to Flip a Prediction}

\author{Jinghan Yang \\
  The University of \\ Hong Kong \\
  \texttt{eciel@connect.hku.hk} \\
  \And
  Linjie Xu \\
  Queen Mary University  \\of London \\
  \texttt{linjie.xu@qmul.ac.uk} \\
  \And
  Lequan Yu  \\
  The University of\\ Hong Kong \\
  \texttt{lqyu@hku.hk} \\}

\begin{document}
\maketitle
\begin{abstract}
When facing an unsatisfactory prediction from a machine learning model, users can be interested in investigating the underlying reasons and exploring the potential for reversing the outcome. 
We ask: 
To flip the prediction on a test point $x_t$, how to identify the smallest training subset $\mathcal{S}_t$ that we need to \textbf{relabel}?
We propose an efficient algorithm to identify and relabel such a subset via an extended influence function for binary classification models with convex loss.
We find that relabeling fewer than 2\% of the training points can always flip a prediction.
This mechanism can serve multiple purposes: (1) providing an approach to challenge a model prediction by altering training points; (2) evaluating model robustness with the cardinality of the subset (i.e., $|\mathcal{S}_t|$); we show that $|\mathcal{S}_t|$ is highly related to the noise ratio in the training set and $|\mathcal{S}_t|$ is correlated with but complementary to predicted probabilities; and (3) revealing training points lead to group attribution bias. 
To the best of our knowledge, we are the first to investigate identifying and relabeling the minimal training subset required to flip a given prediction. \footnote{Code and data to reproduce experiments are available at \url{https://github.com/ecielyang/Relabeling}.}

\end{abstract}

\section{Introduction}

The interpretability of machine learning systems is a crucial research area as it aids in understanding model behavior, facilitating debugging, and enhancing performance \citep{adebayo2020debugging, han2020explaining,pezeshkpour-etal-2022-combining,teso2021interactive, marx2019disentangling}. A common approach involves analyzing the model's predictions by tracing back to the training data \citep{hampel1974influence,cook1980characterizations,cook}. Particularly, when a machine learning model produces an undesirable result, users might be interested in identifying the training points to modify to overturn the outcome. If the identified training points are wrongly labeled, the related determination should be overturned.
{For instance, consider a scenario where a machine learning model evaluates research papers and gives decisions. If an author receives a rejection and disagrees with the result, they might request insight into the specific papers examples used to train the model. If it turns out that correcting a few mislabeled training examples can change the prediction, then the original decision might need reconsideration, possibly accepting the paper instead. This concept is referred to contesting the predictions made by automatic models \citep{hirsch2017designing, vaccaro2019contestability}. When using such models, users should have the right and ability to question and challenge results, especially when these results impact them directly \citep{almada2019human}. Our research is geared towards offering a mechanism for users to challenge these predictions by tracing back to the training data.}

 \begin{figure}[!t]
 \centering
    \includegraphics[scale=0.47]{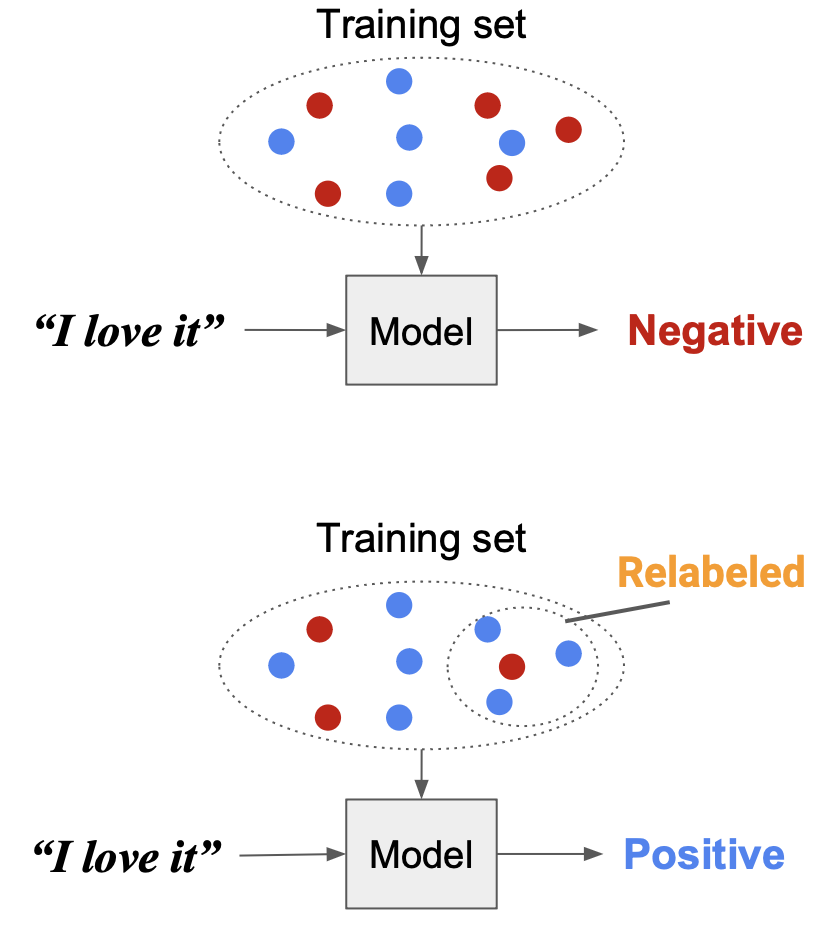}
    \caption{The question we seek to answer is: which is the smallest subset of the training data that needs to be relabeled in order to flip a specific prediction from the model?}
    \label{fig:main}
\end{figure}

\begin{table*}[h]
\centering
\renewcommand{\arraystretch}{1.5}
\scalebox{0.85}{
\begin{tabular}{p{4.3cm} p{1.7cm} p{1.7cm} c p{4cm} p{2.2cm}}
\toprule
\multicolumn{3}{c}{\textbf{Test Point}} & \multirow{2}{*}{\textbf{$|\mathcal{S}_t|$}} & \multicolumn{2}{c}{\textbf{Indentified Training Subset $\mathcal{S}_t$}} \\
\cmidrule{1-3} \cmidrule{5-6}
\textbf{Text} & \textbf{Label} & \textbf{Prediction} & & \textbf{Text} & \textbf{Mislabeled as} \\
\midrule
\emph{The people who can stop it are the ones who pay their wages.} & \textcolor{mygreen}{Non-hate} & \textcolor{red}{Hate} & 1 & \emph{Worker}. & \textcolor{red}{Hate} \\
\midrule
\emph{We will never forget their heroism.} & \textcolor{mygreen}{Non-hate} & \textcolor{red}{Hate} & 1 & \emph{TRUTH NO LIE.} & \textcolor{red}{Hate} \\
\bottomrule
\end{tabular}}
\caption{Examples showcase misclassified test points alongside the identified training set $\mathcal{S}_t$. For each test point, if those training points are relabeled prior to training, the test point can be correctly classified. These training points are intentional noise we manually introduced into the dataset.}
\label{tab:hate-example}
\end{table*}

In this paper, we study the question (visualized in Figure \ref{fig:main}):  
\textit{Given a test point $x_t$ and its associated predicted label $\hat{y}_t$ by a model, how can we find the minimal training subset $\mathcal{S}_t$, if relabeled before training, would lead to a different prediction?} \footnote{We provide a way to investigate the training points instead of retraining the model.}

Identifying $\mathcal{S}_t$ by enumerating all possible subsets of training examples, re-training under each, and then observing the resultant prediction would be inefficient and impractical.
We thus introduce an algorithm for finding such sets efficiently using the extended \emph{influence function}, which allow us to approximate changes in predictions expected as a result of relabeling subsets of training data \cite{koh2019accuracy, flip, kong2021resolving}.

 The identified subset $\mathcal{S}_t$ can be harnessed for a variety of downstream applications.
 Firstly, we discover that $|\mathcal{S}_t|$ can be less than 2\% of the total number of training points, suggesting that relabeling a small fraction of the training data can markedly influence the test prediction. 
Secondly, we observe a correlation between $|\mathcal{S}_t|$ and the noise ratio in the training set. As the noise ratio increases from 0 to 0.5, $|\mathcal{S}_t|$ tends to decrease obviously.
Thirdly, we find that $|\mathcal{S}_t|$ can be small when the model is highly confident in a test prediction, so $|\mathcal{S}_t|$ serves as a measure of robustness that complements to the predicted probability.
Lastly, our approach can shed light on points containing group attribution bias that caused biased determinations. We demonstrate that when such bias exists in the training set,  the corresponding $\mathcal{S}_t$ will significantly overlap with the biased training set.

The contributions of this work are summarized as follows.
(1) We introduce the problem: identifying the minimal subset $\mathcal{S}_t$ of training data, if relabeled,  would result in a different prediction on test point $x_t$; 
(2) We provide a computationally efficient algorithm for binary classification models with convex loss and report performance in text classification problems;
(3) We demonstrate that the size of the subset ($|\mathcal{S}_t|$) can be used to assess the robustness of the model and the training set;
(4) We show that the composition of $\mathcal{S}_t$ can  explain group attribution bias.

\section{Methods}
\label{section:methods} 

This section first demonstrates the algorithm to find the minimal relabel set and shows a case to use the algorithm to challenge the model's prediction.

\subsection{Algorithm}

Consider a binary classification problem with a training dataset denoted as \( \mathcal{Z}^{\text{tr}} = \{z_1, \ldots, z_N\} \). Each data point \( z_i = (x_i, y_i) \) consists of features \( x_i \in \mathcal{X} \) and a label \( y_i \in \mathcal{Y} \). We train a classification model \( f_{w}: \mathcal{X} \rightarrow \mathcal{Y} \), where \( f \) is parameterized by a parameter vector \( w \in \mathbb{R}^p\). By minimizing the empirical risk, this process yields the estimated parameter \( \hat{w} \), defined by:

\begin{equation}~\label{eq1}
  \begin{aligned}
  \hat{w} & := \underset{w}{\text{argmin}} \, \mathcal{R}(w) \\
& = \underset{w}{\text{argmin}} \left( \frac{1}{N} \sum_{i=1}^N \mathcal{L}(z_i, w) + \frac{\lambda}{2} \|w\|^2 \right)
  \end{aligned}
\end{equation}

\( \mathcal{L}(z_i, w) \) represents the loss function that measures the prediction error for a single data point \( z_i \) given the parameters \( w \), and \( \mathcal{R}(w) \) denotes the total empirical risk, which includes a regularization term controlled by the hyperparameter \( \lambda \). We assume that $\mathcal{R}$ is twice-differentiable and strongly convex in $w$, with the Hessian matrix $H_{\hat{w}} := \nabla_{w}^2 \mathcal{R}(\hat{w}) = \frac{1}{N} \sum_{i=1}^N \nabla_{w}^2  \mathcal{L}(z_i, \hat{w}) + {\lambda} \mathit{I}$.

Suppose we relabel a subset of the training points \( \mathcal{S} \subset \mathcal{Z}^{\text{tr}} \) by changing \( y_i \) to \( y_i' \) for each \( (x_i, y_i) \in \mathcal{S} \) and then re-estimate \( w \) to minimize \( \mathcal{R}(w) \), resulting in new parameters \( \hat{w}_{\mathcal{S}} \):

\begin{equation}
\hat{w}_{\mathcal{S}} = \underset{w}{\text{argmin}} \left( \mathcal{R}(w) + \frac{1}{N} \sum_{(x_i, y_i) \in \mathcal{S}} \ell \right)
\end{equation}

where \( \ell = -\mathcal{L}(x_i, y_i, w) + \mathcal{L}(x_i, y_i', w) \) represents the adjustment to the original loss due to the relabeling of points in \( \mathcal{S} \).
 
Due to the large number of possible subsets in the training set, it is computationally impractical to relabel and retrain models for each subset to observe prediction changes.
\citet{warnecke2021machine, kong2021resolving} derived the influence exerted by relabeling a training set $\mathcal{S}$ on the \emph{loss} incurred for a test point $t$ as: 
\begin{equation}
\nabla_w \mathcal{L}(z_t, \hat{w})^\intercal {\Delta_i w},
\label{eq:inf-classic-loss}
\end{equation}
where $\Delta_i w = \frac{1}{N} H^{-1}_{\hat{w}} \sum_{(x_i, y_i)\in S}\nabla_{w} \ell$
is the change of parameters after relabeling training points in $\mathcal{S}$.
Instead, we estimate the influence on \emph{predicted probability} result by relabeling the training subset $\mathcal{S}$ as:
\begin{equation}
    \Delta_{t}f := \nabla_{w} f_{\hat{w}}(x_t)^\intercal \Delta_i w,
    \label{eq:inf-IP}
\end{equation}
which is named as {\bf IP-relabel}. 

{Based on this IP-relabel and adopting the algorithm proposed by \citet{broderick2020automatic, yang2023many}, we propose the Algorithm~\ref{alg:alg1} to find a training subset $\mathcal{S}_t$ to relabel, which would result in flipping the test prediction $\hat{y}_t$ on $x_t$.
Our approach initiates by approximating the change in predicted probability $\Delta_t f$ for a test point $x_t$, which results from the relabeling of each training point.
Subsequently, we iterate through all the training points in a descending order of their influence from the most decisive to the least. During each iteration, we accumulate the change in predicted probability $\Delta_t f$.
When the cumulative change causes the output $\hat{y}_t$ to cross a predefined threshold, the algorithm identifies $\mathcal{S}_t$. If, however, the output fails to cross the threshold even after examining the entire training set, the algorithm is unable to find the set $\mathcal{S}_t$. 
{For \(N\) training points and the parameter \(w\) in \(\mathbb{R}^p\), our algorithm requires \(O(p^3)\) to compute the inverse of the Hessian matrix for the total loss and \(O(N p^2)\) to calculate the IP-relabel for each training point. Therefore, the overall computational complexity is \(O(p^3 + N p^2)\). We also include the running time of our experiments in Appendix \ref{app:running-time}}.

\newcommand{\mycomment}[1]{}

\subsection{Case Study}
{
In this section, we present an example to demonstrate how our method can be used to challenge the predictions of machine learning models. We employ the Hate Speech dataset \citep{gibert2018hate}, which encompasses instances of hate communication that target specific groups based on characteristics such as race, color, ethnicity, etc. On social media platforms, users found engaging in hate speech are typically banned.}

{We implement a linear regression model to classify hate speech on the internet. We intentionally introduced noise into the training dataset by mislabeling 1,000 data points (out of 9632, switching labels from 1 to 0 and vice versa). This deliberate noise in the training set can result in additional misclassifications during model testing.}

{As demonstrated in Table \ref{tab:hate-example}, for each test instance, Algorithm \ref{alg:alg1} pinpoints the specific training data points that, when relabeled before training, could change the prediction of the test point. The table showcases two instances where the model misclassified test points. The corresponding training sets, $\mathcal{S}_t$, consist of training points that closely resemble the test cases but were erroneously labeled. Given that the classifications can be altered by relabeling the small subset of mislabeled training data, determinations based on these classifications, such as banning users, warrant careful reconsideration.}

\begin{algorithm}[!ht]
\caption{An algorithm to find a minimal subset to flip a test prediction}
	\label{alg:alg1}
\DontPrintSemicolon
  \KwInput{
  $f$: Model;
  $\mathcal{Z}^{\text{tr}}$: Full training set; 
  $N$: number of total training points;
  $\mathcal{Z}^{\text{tr}’}$: Relabeled full training set; 
  $\hat{w}$: Parameters estimated $\mathcal{Z}^{\text{tr}}$;  
  $\mathcal{L}$: Loss function; 
  $x_t$: A test point; 
  $\tau$: Classification threshold (e.g., 0.5)}
  \KwOutput{$\mathcal{S}_t$: minimal train subset identified to flip the prediction ($\emptyset$ if unsuccessful)}
  $H \leftarrow \nabla_{w}^2  \mathcal{L}(\mathcal{Z}^{\text{tr}}, \hat{w})$ \\ 
  $\nabla_w l \leftarrow -\nabla_{w} \mathcal{L}(\mathcal{Z}^{\text{tr}}, \hat{w}) +\nabla_{w} \mathcal{L}(\mathcal{Z}^{\text{tr}'}, \hat{w})$ \\
  $\Delta w \leftarrow \frac{1}{N} H^{-1} \nabla_w l$ \\ 
  $\Delta_t f \leftarrow \nabla_{w} f_{\hat{w}}(x_{t})^\intercal \Delta w$\\
  $\hat{y}_t \leftarrow f(x_{t}) > \tau$ \tcp*{Binary prediction}
    \tcp{Sort instances (and estimated output differences) in order of the current prediction} 
  ${\tt direction} \leftarrow \{\uparrow$ if $\hat{y}_t$ else $\downarrow$\} \\
  ${\tt indices} \leftarrow {\tt argsort}(\Delta_t f, {\tt direction})$\\
  $\Delta_t f \leftarrow {\text{\tt sort}}(\Delta_t f, {\tt direction})$ \\

   	\For{$k=1$ ... $|\mathcal{Z}^{\text{tr}}|$} 
   	{
   	$\hat{y}_t' = (f(x_{t}) + {\text{\tt sum}}(\Delta_t f [:k])) > \tau$ 
   	\If{$\hat{y}'_t \neq \hat{y}_t$}
        {
            \KwRet $\mathcal{Z}^{\text{tr}}[{\tt indices}[:k]]$
        }
   	}
   	\KwRet $\emptyset$
\end{algorithm}

\section{Experiments}

{We provide an overview of our experiments:}
\begin{enumerate}
    \item We introduce our experimental setup and then validate Algorithm \ref{alg:alg1} in Sec \ref{sec:exp-setting} and \ref{sec:exp-vali}. Our results confirm that we can effectively change the test predictions by relabeling revealed points and subsequent model retraining.
    \item Sec \ref{sec:exp-size} analyzes the magnitude of $|\mathcal{S}_t|$ across various datasets and models, emphasizing its correlation with predicted probability and noise ratio. This showcases its utility in analyzing the robustness of training points and models.
    \item We further delve into the integration of subset $\mathcal{S}_t$ in Sec \ref{sec:exp-inte}, demonstrating its potential to highlight biased training data.
    \item In Sec \ref{sec:exp-compare},  we compare our method against other methods to alter training points to flip test prediction, illustrating that our method revealed a smaller training subset.
\end{enumerate}

\subsection{Experimental Setting}
\label{sec:exp-setting}

\vspace{0.25em}
\noindent{\bf Datasets.} We use a tabular dataset: Loan default classification \cite{loan_data}, and text datasets: Movie review sentiment 
\cite{socher2013recursive}; Essay grading \cite{essay_data};  Hate speech \cite{gibert2018hate}; and Twitter sentiment \cite{go2009twitter} to evaluate our method.

\vspace{0.25em}
\noindent{\bf Models.} We consider the $l_2$ regularized logistic regression to fit the assumption on influence function.
As features, we consider both bag-of-words and neural embeddings induced via BERT  \cite{devlin2018bert} for text datasets. We report basic statistics describing our datasets and model performance in Section \ref{app:data-model-details}.

\begin{table}
\small
\centering 
\scalebox{0.76}{
\begin{tabular}{l l c c c}
\toprule
{\textbf{Dataset}} & {\textbf{Features}} & {\textbf{Found $\mathcal{S}_t$}} & {\textbf{Flip Successful}} & {\textbf{Successful Ratio}} \\
\hline 
\emph{Loan} &BoW & 61\% & 49\% & 80\% \\ \cmidrule(lr){2-5}
\emph{Movie} & BoW & 100\% & 72\% & 72\%\\
\emph{reviews} & BERT & 100\% & 73\% & 73\%\\ \cmidrule(lr){2-5}
\emph{Essays} & BoW & 77\% & 40\% & 52\%\\
 & BERT &  76\% & 39\% & 51\%\\ \cmidrule(lr){2-5}
\emph{Hate} & BoW          &  99\% & 87\% & 87\%   \\ 
\emph{speech}& BERT &  99\% & 86\% & 87\%  \\ \cmidrule(lr){2-5}
\emph{Tweet} & BoW         & 100\% & 75\% & 75\% \\
\emph{sentiment} & BERT        & 100\% & 68\% & 68\% \\
\bottomrule
\end{tabular}}
\caption{Percentages of text examples for which Algorithm \ref{alg:alg1} successfully identified a set $\mathcal{S}_t$ (2nd column) and for which upon flipping these instances and retraining the prediction indeed flipped (3rd column). The "Successful Ratio" is obtained by divide the percentages in the "Flip Successful" column by those in the "Found $\mathcal{S}_t$" column.}
\label{table:k-flips}
\end{table}

\subsection{Algorithm Validation}
\label{sec:exp-vali}

\vspace{0.25em}
\noindent{\bf
How effective is our algorithm at finding $\mathcal{S}_t$ and flipping the corresponding prediction?} 
As shown in Table~\ref{table:k-flips}, the frequency of finding $\mathcal{S}_t$ varies greatly among datasets. 
For the movie reviews and tweet datasets, Algorithm~\ref{alg:alg1} returns a set $\mathcal{S}_t$ for approximately 100\% of test points. On the other hand, for the simpler loan data, it only returns $\mathcal{S}_t$ for approximately 60\% of instances. Results for other datasets fall between these two extremes.
When the algorithm successfully finds a set $\mathcal{S}_t$, relabeling all $(x_i, y_i) \in \mathcal{S}_t$ almost enables the re-trained model to flip the prediction $\hat{y}_t$ (as indicated in the right-most column of Table \ref{table:k-flips}). 

\vspace{0.25em}
\noindent{\bf Comparison with other methods.}
We draw comparisons between IP-relabel and several other methods \citep{pezeshkpour2021empirical}, including IP-remove \citep{yang2023many}, influence function \citep{IF}, and three gradient-based instance attribution methods on a logistic regression model to the movie review dataset \citep{RIF, GC}:

\noindent{1. $RIF = cos(H^{-\frac{1}{2}} \nabla_{w}\mathcal{L}(x_t), H^{-\frac{1}{2}} \nabla_{w}\mathcal{L}(x_i))$}

\noindent{2. $GD = \left< \nabla_{w}\mathcal{L}(x_t), \nabla_{w}\mathcal{L}(x_i) \right>$}

\noindent{3. $GC = cos( \nabla_{w}\mathcal{L}(x_t), \nabla_{w}\mathcal{L}(x_i))$
}

\noindent{We also randomly select subsets of training data and relabel them.
We graph the average change in predicted probability for 100 randomly chosen test points in Figure \ref{fig:change_prob}. These probabilities are from the model trained before and after relabeling the top $k$ training points ranked on the scores above. 
Our analysis indicates that IP-relabel shows a more significant impact in the test predicted probability compared to the impact of removing training points as ranked by other methods.}

 \begin{figure*}[h]
    \centering
    \includegraphics[scale=0.45]{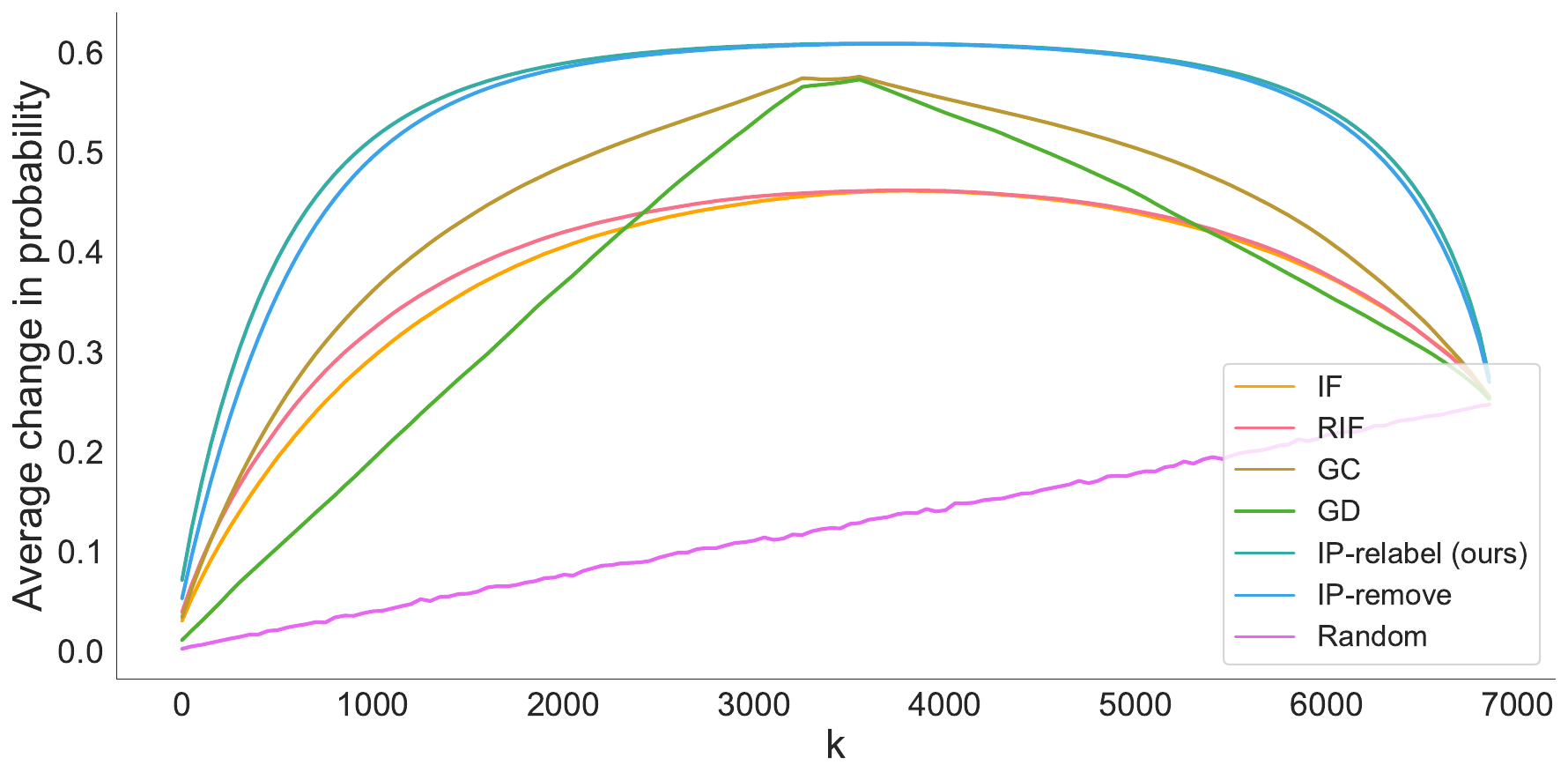}
    \caption{The relationship between the average of absolute difference on predicted probabilities for sampled test points results from relabeled $k=|\mathcal{S}_t|$ training points, using different methods on movie review dataset.}
    \label{fig:change_prob}
\end{figure*}

\vspace{0.25em}
\noindent{\bf Running time of Algorithm \ref{alg:alg1}.} We recorded the average running time of Algorithm \ref{alg:alg1} to find $\mathcal{S}_t$ for test points in different datasets in Table \ref{tab:run-time} on Apple M1 Pro CPUs. For one test point, it just takes milliseconds to go through the whole training set (the training set sizes are provided in \ref{app:data-model-details}) to find $\mathcal{S}_t$. 

\begin{table}[h]

\centering
\begin{tabular}{ccc}
\hline
\textbf{Dataset} & \textbf{BoW (ms)} & \textbf{BERT (ms) } \\
\hline
Movie Reviews & 19.04 & 140.51 \\
Essays & 160.01 & 265.09 \\
Hate speech & 103.70 & 299.46 \\

Tweet & 58.42 & 260.75 \\

Loan & 63.97 & / \\
\hline
\end{tabular}
\caption{Average running time (in milliseconds) of Algorithm \ref{alg:alg1} to find $\mathcal{S}_t$ for a test point in different datasets.}
\label{tab:run-time}
\end{table}

\subsection{$|\mathcal{S}_t|$ Quantifies Model Robustness}
\label{sec:exp-size}

\begin{figure}[h!]
    \centering
    \includegraphics[scale=0.25]{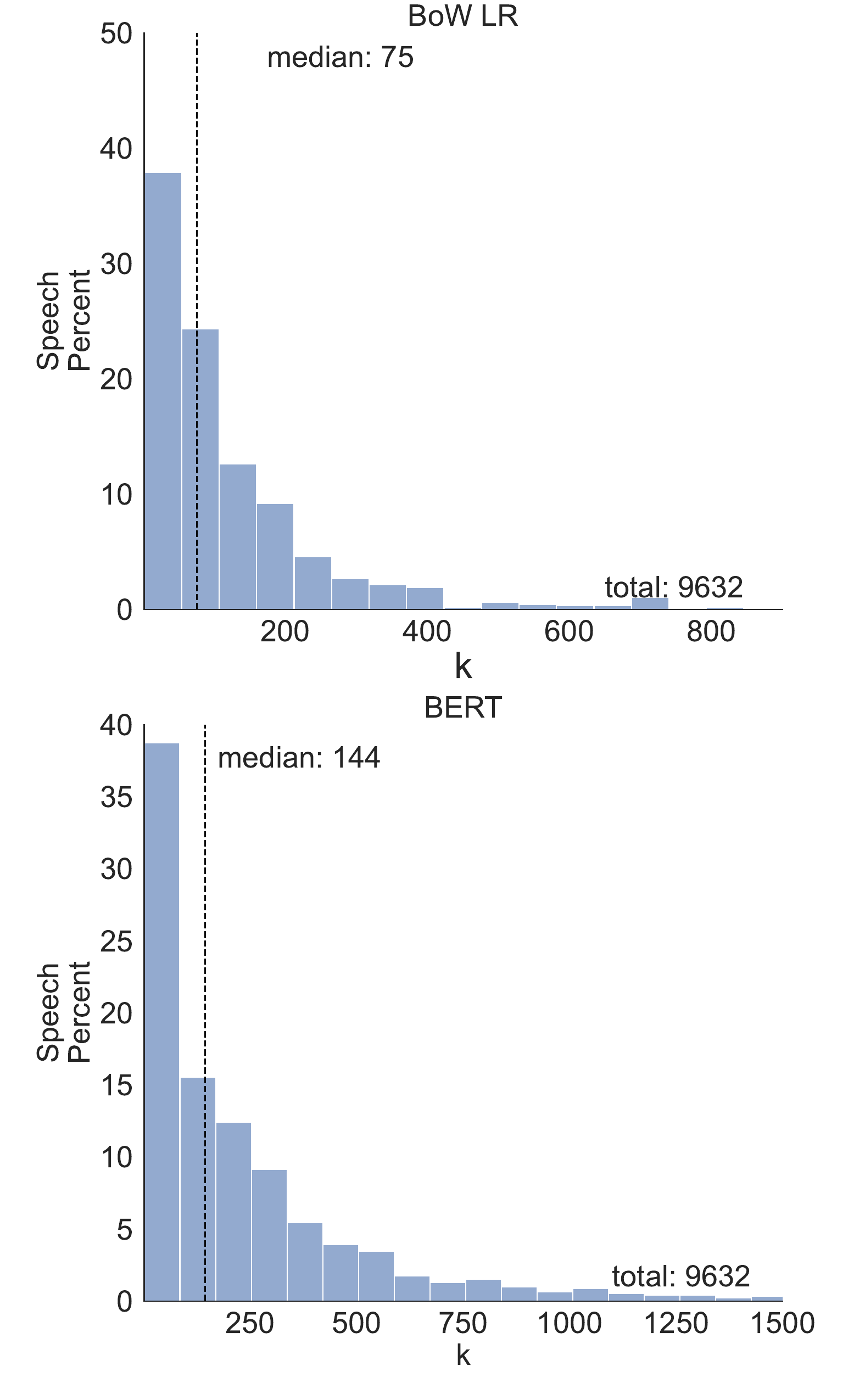}
    \caption{
    The histogram shows the distribution of $k=|\mathcal{S}_t|$ on the hate speech dataset, i.e. the minimal number of points that need to be relabeled from the training data to change the prediction  $\hat{y}_t$ of a specific test example $x_t$.}
    \label{fig:dist-k}
\end{figure}

\noindent{\bf Relabeling less than 2\% training data can usually flip a prediction.} 
The empirical distributions of $k$ values for subsets $\mathcal{S}_t$ identified by Algorithm~\ref{alg:alg1} can be seen in Figure \ref{fig:dist-k} for the representative hate speech datasets (full results are in the Appendix). 
The key observation is that when $\mathcal{S}_t$ is found, its size is often relatively small compared to the total number of training instances. 
In fact, for many test points, relabeling less than 2\% instances would have resulted in a flipped prediction.

\vspace{0.5em}
\noindent{\bf BERT demonstrates greater robustness than LR based on $|\mathcal{S}_t|$ measures.}
{
For a proficiently trained model, relabeling a larger subset of training data in order to alter a correct test prediction suggests greater model robustness. In Figure \ref{fig: size-bert-lr}, we present a comparison of the average values of $|\mathcal{S}_t|$ for common test data points where both BERT and LR model predictions were successfully altered using our method. The results indicate that BERT typically demands the relabeling of more training data points than the LR models do. This observation supports the utility of our method in gauging the relative robustness of different models.
}

\begin{figure}
    \centering
    \hspace*{-0.7cm}
    \includegraphics[scale=0.3]{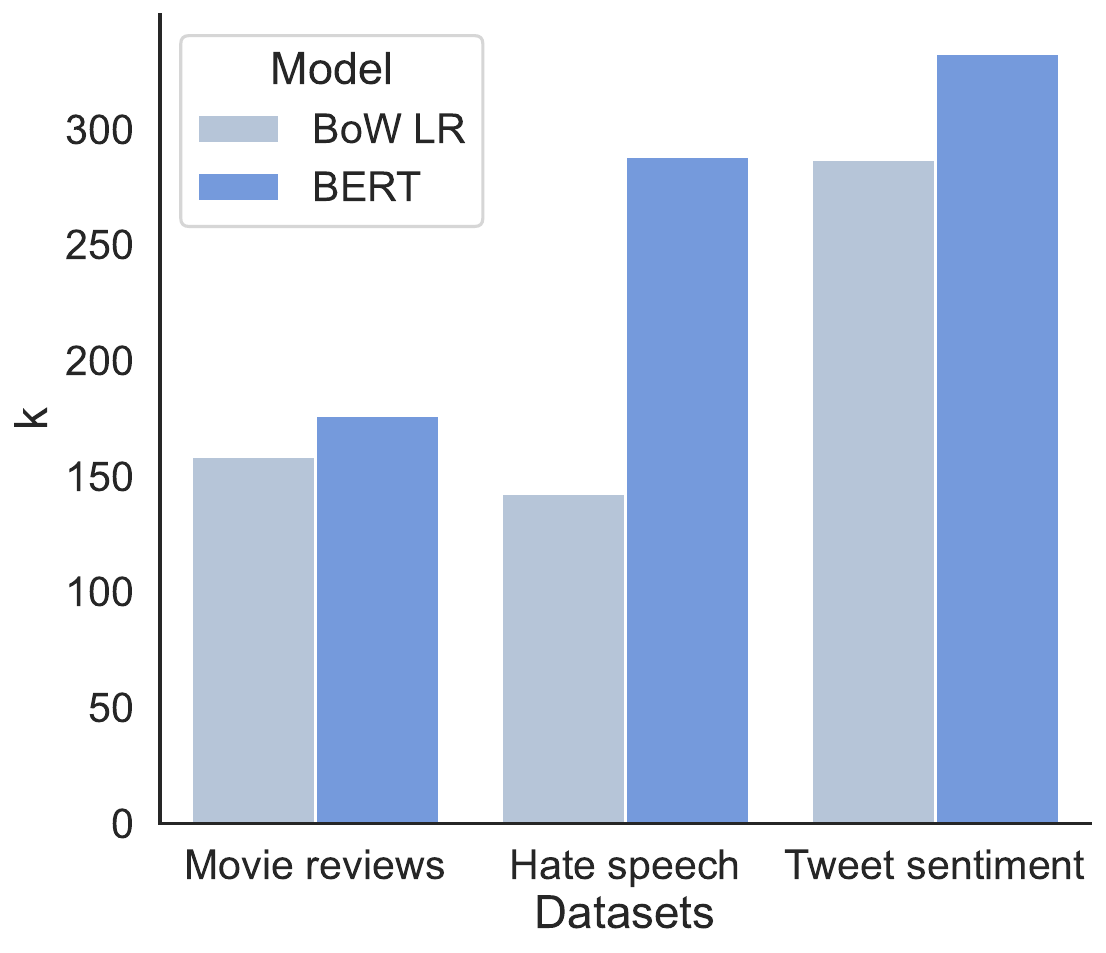}
    \caption{Comparison of the average $k = |\mathcal{S}_t|$ values for shared test points under both BERT and LR models that were successfully flipped by our method.}
    \label{fig: size-bert-lr}
\end{figure}

 \vspace{0.5em}
\noindent {\bf Correlation between $k$ and the predicted probability.}
Does the size of $\mathcal{S}_t$ tell us anything beyond what we might infer from the predicted probability $p(y_t=1)$? 
In Fig \ref{fig:kp} we show a scatter of $k=|\mathcal{S}_t|$ against the distance of the predicted probability from 0.5 on speech dataset. 
There are test instances of the model being confident, but relabeling a small set of training instances would overturn the prediction.
In Sec \ref{sec-full-plots}, there are datasets where the $k$ can be highly correlated with probability.

\begin{figure}[h!]
    \centering
    \includegraphics[scale=0.3]{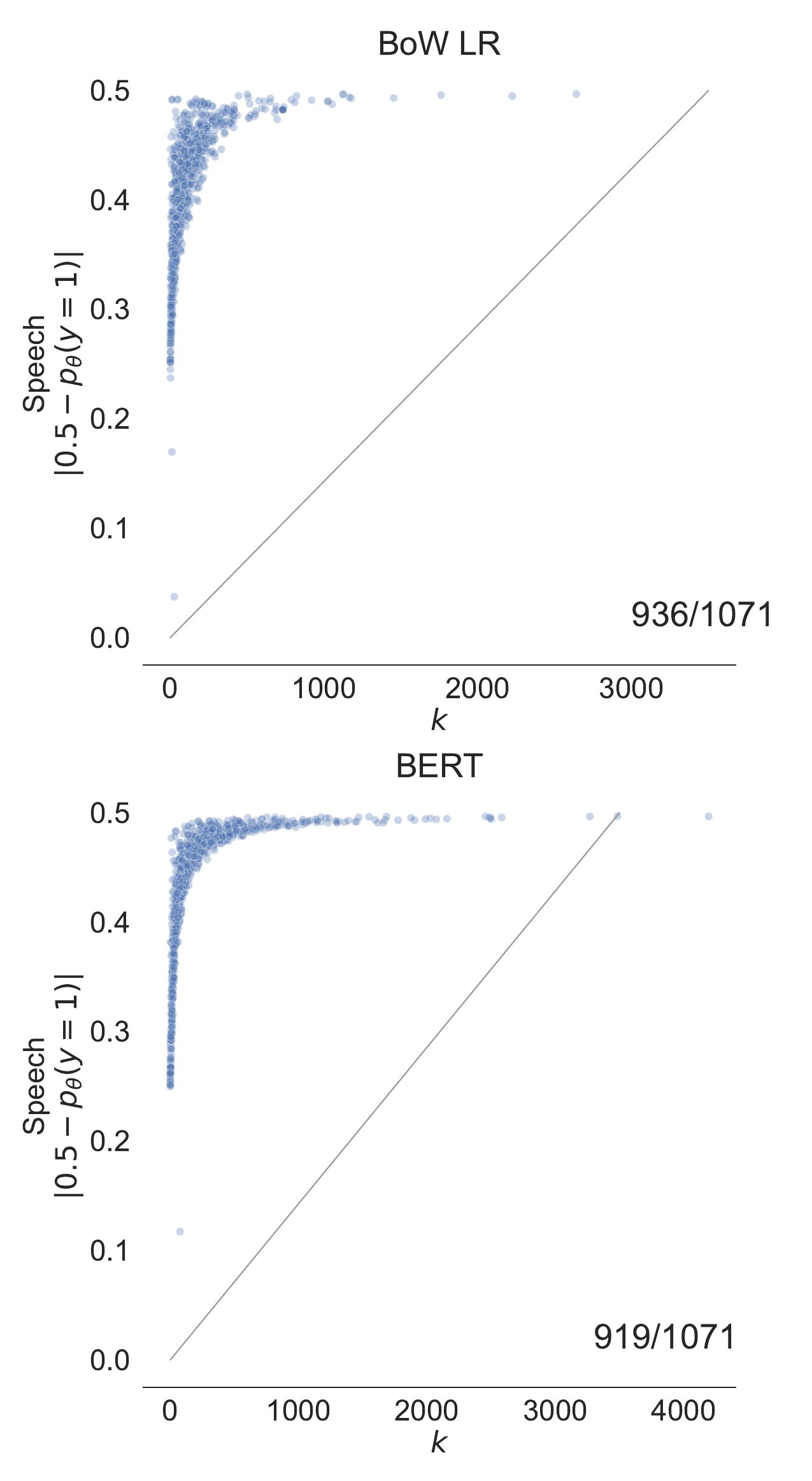}
    \caption{
    The correlation between the predicted probabilities of certain test examples and $k=|\mathcal{S}_t|$ on the hate speech dataset. 
    For test examples where the model is highly certain about its prediction, the prediction can be flipped  by relabeling a small number of data points from the training set.}
    \label{fig:kp}
\end{figure}

 \vspace{0.5em}
\noindent{\bf How is $|\mathcal{S}_t|$ correlated with the noise ratio?}
Figure \ref{fig:noise-ratio} shows how $|\mathcal{S}_t|$ and the model's accuracy vary when we increase the noise ratio from 0 to 0.9. We introduce noise to the training set by incrementally relabeling a portion of training points, from 0 to 0.9 in steps of 0.1. When the noise ratio increases from 0 to 0.5, we observe a decline in $|\mathcal{S}_t|$. However, as the noise ratio rises from 0.5 to 0.9, $|\mathcal{S}_t|$ starts to increase.
Interestingly, within the noise ratio interval of 0 to 0.3, the model's accuracy does not demonstrate a noticeable decline. This suggests that $|\mathcal{S}_t|$ can be an additional metric for assessing the model's robustness complementary to accuracy under different noise ratios.

\begin{figure*}
    \centering
    \includegraphics[scale=0.40]{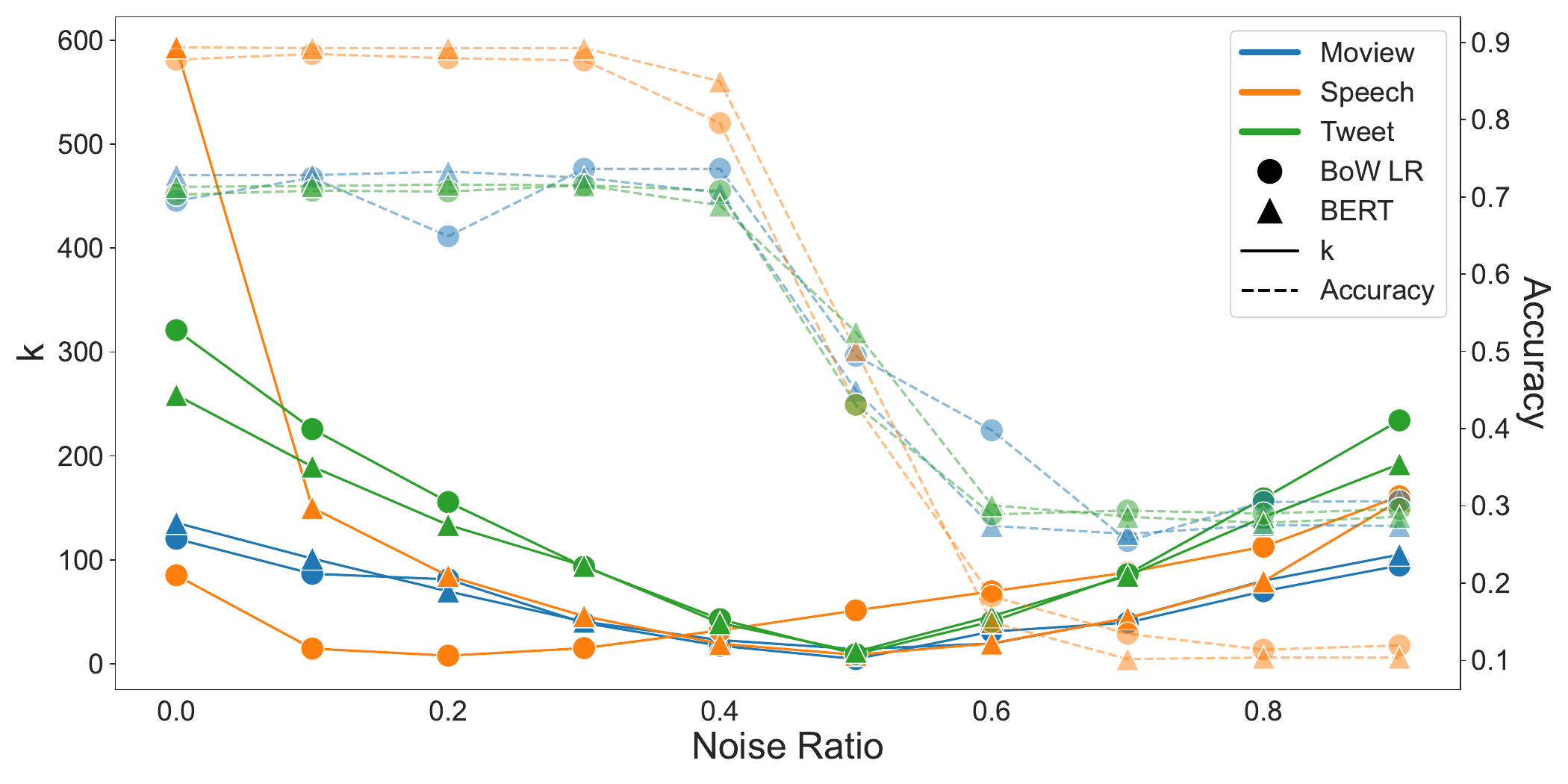}
    \caption{Average of $k=|\mathcal{S}_t|$ (solid line) and model's accuracy (dashed line) for the test dataset with noise ratio from 0 to 0.9. When the noise ratio increases from 0 to 0.3, $k$ decreases apparently, while the model's accuracy does not demonstrate a noticeable decline.}
    \label{fig:noise-ratio}
\end{figure*}

\subsection{Composition of $\mathcal{S}_t$ Contributes Bias Explanation} \label{sec:exp-inte}
 \vspace{0.5em}
Group attribution bias in machine learning refers to a model's inclination to link specific attributes to a particular group, potentially resulting in biased predictions. We show that the integration of $\mathcal{S}_t$ is associated with group attribution biased in training data.
As a case, we manually introduce group attribution bias into the loan default dataset \cite{loan_data}, designed to predict potential defaulters for a consumer loan product.
We augment a dataset containing basic consumer features with a manually added discrete "tag" feature, arbitrarily assigning 40\% as "tag $X$" and 60\% as "tag $Y$" We then introduce bias by relabeling 90\% of the qualified "tag $X$" as "default." This biased set is defined as $\mathcal{B}$, where the wrong label tightly links with the feature "tag $X$." A logistic regression model is subsequently trained with this modified dataset.

We apply Algorithm \ref{alg:alg1}  to misclassified test points and compute the proportion in each resulting subset $\mathcal{S}_t$ belonging to $\mathcal{B}$. The average proportions are 60\% for "tag $X$" and 23\% for "tag $Y$" misclassified data.
The higher proportion in "tag $X$" suggests that the misclassification of eligible "tag $X$" individuals mainly results from the biased training set $\mathcal{B}$, whereas for "tag $Y$" individuals may be due to other reasons like model oversimplification. Thus, our approach can highlight training points contributing to group attribution bias.

\subsection{Comparison between Removal and Relabeling}
\label{sec:exp-compare}

\begin{table*}[th]
\centering
\begin{tabular}{lcccccc}
\toprule
& \multicolumn{3}{c}{\textbf{Noisy points in $\mathcal{S}_{t1}$}} & \multicolumn{3}{c}{\textbf{Normal points in $\mathcal{S}_{t2}$}} \\ 
\cmidrule(lr){2-4} \cmidrule(lr){5-7}
& \textbf{Loan} & \textbf{Movie reviews} & \textbf{Speech} & \textbf{Loan} & \textbf{Movie reviews} & \textbf{Speech} \\
\hline
\textbf{Removal Alg1} & 47.9 & 1.8 & 146.8 & 30.6 & 2.1 & 31.9 \\
\textbf{Removal Alg2} & 45.6 & 1.8 & 104.2 & 27.0 & 2.1 & 21.0 \\
\textbf{Relabeling (ours)} & {\bf 11.6} & {\bf 0.8} & {\bf 55.8} & {\bf 22.9} & {\bf 1.3} & {\bf 8.2} \\
\bottomrule
\end{tabular}
\caption{Average number of points to relabel and remove to flip a test prediction, categorized by noisy and normal points. Relabeling consistently leads to smaller sets of both noisy and normal points being altered.}
\label{table:compare-noise-tmp}
\end{table*}

{In this section, we compare two ways to alter training points such that the alternation can result in the flipping of a test point: relabeling and removal. We show that the relabeling mechanism can reveal a smaller training  subset, thus saving the cost of investigating suspicious training points.}

{\citet{kong2021resolving} firstly propose an algorithm to find the training subset to remove to flip a test prediction for economy models, which we denote as "Removal Alg1" in Table \ref{table:comapre-remove-label}.
\citet{yang2023many} employ the same algorithm on machine learning models and improve it to return a smaller training set, denoted as "Removal Alg2".}

We aim to show that when noise is present in the training set, the relabeling mechanism consistently uncovers a smaller subset of influential points from the noisy training set while affecting fewer standard points. 
To demonstrate this, we introduced a 30\% noise factor into the training set by flipping labels of normal points, denoted as $\mathcal{N}$, which increased misclassified test points. We identified the training set $\mathcal{S}_t$ using the three methods for these misclassified test points.
We divided the identified training points $\mathcal{S}_t$ into two categories: training points belonging to the noise set $\mathcal{S}_{t1} = \mathcal{S}_t \cap \mathcal{N}$, and those that do not belong to the noise set $\mathcal{S}_{t2} = \mathcal{S}_t \setminus \mathcal{N}$.
The results presented in Table \ref{table:compare-noise-tmp} demonstrate that both the $S_1$ and $S_2$ subsets identified through the relabeling process are smaller than those identified through removal. This suggests that considering relabeling training points can more effectively discern fewer noisy and regular training points, saving the cost to investigate more suspicious points. We also show the conclusion holds when there is no noise in the training set in Sec \ref{no-noise-remove-relabel}.

\section{Related Work}
\label{section:related-work}

\noindent{\bf The holding of model predictions.}
Several studies have explored the changes of a model behavior and its factors. \citet{pmlr-v162-ilyas22a} analyzed model behavior changes based on different training data. 
\citet{harzli2022minimal} studied the change of a specific prediction by finding a smallest informative feature set to analize economy models. 
Additionally, research on \emph{counterfactual examples}  aims to explain predicted outcomes by identifying the feature values that caused the given prediction \cite{kaushik2019learning}.
Recent studies  investigated the influence function in machine learning to answer the question of "How many and which training points need to be removed to alter a specific prediction?" \citep{broderick2020automatic, yang2023many}. We follow these two works and propose an alternative way to alter the training points by asking, "How many and which training points would need to be relabeled to change this prediction?"

 \vspace{0.5em}
\noindent{\bf Trustworthy machine learning} is important in today's era, given the pervasive adoption of artificial intelligence systems in our everyday lives.
Previous work emphasizes contestability as a key facet of trustworthiness, advocating for individuals' right to challenge AI predictions \citep{vaccaro2019contestability,almada2019human}.
This may involve providing evidence or alternative perspectives to challenge AI-derived conclusions \citep{hirsch2017designing}.
Our mechanism offers a way to draw upon training data as evidence when contest AI determination.
In line with advancing model fairness, it's crucial to address training data related to noise \citep{wang2018devil,kuznetsova2020open} and biases \citep{osoba2017intelligence, howard2018ugly}. Our research shows that, despite different noise ratios, the model's accuracy remains relatively consistent, yet there is a significant variation in the size of the subset $\mathcal{S}_t$. Furthermore, we demonstrate that in scenarios where group attribution bias is present, our method can aid in identifying the associated training points.

 \vspace{0.5em}
\noindent{\bf Influence function} offers tools for identifying training data most responsible for a particular test prediction \citep{hampel1974influence,cook1980characterizations,cook}.
By uncovering mislabeled training points and/or outliers, influence can be used to debug training data and provide insight for the result generated by neural networks \citep{IF,adebayo2020debugging, han2020explaining,pezeshkpour-etal-2022-combining,teso2021interactive}.
\citet{warnecke2021machine} extend influence function to measure the influence of alternation in training points' feature and label and apply it to machine unlearning. 
Furthermore, \citet{kong2021resolving} also extended influence on the effect of relabeling training points but utilized this measure to identify and recycle noisy training samples, leading to enhanced model performance at the training stage.
Our research emphasizes utilizing this measure to determine which training subsets should be relabeled to question machine learning model predictions, and we delve into the factors influencing the integration and size of the identified subsets.

\section{Discussion and Future Work}

\noindent\textbf{{Extend the method to complex models.}}
{In today's landscape dominated by large language models (LLMs), researchers are trying to integrate machine learning models into various decision-making processes, ranging from medical diagnoses \citep{shaib2023summarizing} to legal judgments \citep{jiang2023legal} and academic paper reviews  \citep{liang2023large}.
However, LLMs are black-box models and hard to explain despite their immense capabilities. They are prone to challenges including, but not limited to, social biases \citep{hutchinson2020social, bender2021dangers, abid2021persistent, weidinger2021ethical, bommasani2022opportunities} and the spread of misinformation \citep{evans2021truthful, lin2022truthfulqa}. These immediate issues might be precursors to more profound, long-term risks for making decisions based on AI systems.}

{As we harness these models to make critical decisions, it becomes imperative to delve into the root causes of any erroneous determinations. As outlined in our research, our proposed method offers a pathway to trace the origins of such errors back to specific training data points. As the first to state this problem, we primarily focus on linear regression and BERT with a classifier.
In the future, we envision our methodology applying to even more complex models. A recent study extends the influence function to LLMs to understand how training data alterations can impact model predictions \citep{grosse2023studying}. Building upon this foundation, adapting our approach for LLMs is promising for future exploration.}
{Because IP-relabel calculates how the predicted probability changes when training points are relabeled, we can readily adapt our method for multi-class tasks. If we know the desired label to which we want to change certain training points, we can simply adjust the threshold in Algorithm 1 to alter the test predictions accordingly.}

\vspace{0.25em}
\noindent\textbf{{Improve model performance.}}
{Instead of scaling up the number of datasets, we can focus on current data and alter them to improve the quality, enhancing downstream performance, as suggested by the reviewer.
For instance, \citet{kong2021resolving} introduced a framework for relabeling incoming training points that may contain noise. This approach successfully improved the model's performance on test data.
Similarly, \citet{teso2021interactive} developed an algorithm to identify and eliminate potentially noisy training points, thereby improving the overall quality of the training set and, consequently, the model's performance. Both studies utilized influence functions, a concept we employ, albeit with a distinct formulation as indicated in Equ.~\eqref{eq:inf-IP}.
Similarly, future work can consider enhancing the overall model performance by improving the data quality through identifying and relabeling training points that can flip wrong test predictions.
}

\section{Conclusions}
In this work, we introduce the problem of identifying a minimal subset of training data, $\mathcal{S}_t$, which, if relabeled before training, would result in a different test prediction. 
{We propose a computationally efficient algorithm to address this task and evaluate its performance within binary classification models with convex loss.} 
In the experiment, we illustrate that the size of the subset $|\mathcal{S}_t|$ can serve as a measure of the model and the training set's robustness. Lastly, we indicate that the composition of $\mathcal{S}_t$ can reveal training points that cause group attribution bias.

\section{Limitations and Risks}
In our study, we've extensively used influence functions to solve the problem. However, being aware of fundamental limitations is crucial: they tend to be only effective in convex loss. 
The overarching goal of pinpointing a minimal subset within the training data, such that a change in labels leads to a reversal in prediction, isn't exclusively achievable via approximations rooted in influence functions. This approach is favored in our work due to its intuitive nature and wide use.
In addition, while Algorithm~\ref{alg:alg1} currently shows less than optimal performance on the essay dataset, this presents an opportunity for further investigation. Specific characteristics unique to this dataset might influence the performance, opening up a valuable avenue for future research. 

There exists an inherent risk wherein the same approach could be exploited to engender biased determinations. Specifically, by intentionally mislabeling genuine training data and subsequently retraining the model, actors with malicious intent might be able to invert just determinations, thereby compromising the model's integrity and fairness. To counteract this risk, strategies such as regular data integrity checks, stringent access control, and employing model robustness techniques can be integrated, thereby ensuring the preservation of model authenticity and shielding against adversarial exploits.

\section*{Acknowledgements}
We are thankful to the reviewers for their thoughtful and helpful advice.

\bibliography{anthology,custom}
\bibliographystyle{acl_natbib}

\appendix

\section{Appendix}
\label{sec:appendix}

\subsection{Datasets and model details}
\label{app:data-model-details}
We present basic statistics describing our text classification datasets in Table \ref{table:dataset-info}. 
We set the threshold for the hate speech data as 0.25 ($\tau=0.25$) to maximize the F1 score on the training set. For other datasets, we set the threshold as 0.5.
For reference, we also report the hyperparameters and predictive performance realized by the models considered on the test sets of datasets in Table \ref{table:model-performance}.

\begin{table}
\small
\centering 
\begin{tabular}{l l l l}
\hline
{\textbf{Dataset}} & {\textbf{\texttt{\#} Train}} & {\textbf{ \texttt{\#} Test}} & {\textbf{\% Pos}} \\
\hline 
Loan                 & 21120 & 2800 & 0.50   \\
Movie reviews          & 6920 & 872 & 0.52  \\
Essay                  & 11678 & 1298 & 0.10 \\
Hate speech            & 9632 & 1071 & 0.11   \\
Tweet sentiment        & 18000 & 1000 & 0.50  \\
\hline
\end{tabular}
\caption{Dataset information.}
\label{table:dataset-info}
\end{table}

\begin{table}
\small
\centering 
\begin{tabular}{l l l l l}
\hline
{\textbf{Models}} & {\textbf{Accuracy}} & {\textbf{F1-score}} & {\textbf{AUC}} & {\textbf{l2}} \\
\hline 
 \multicolumn{3}{c}{\emph{Loan}} \\
LR & 0.79 & 0.80 & 0.88 & 100\\
 \multicolumn{3}{c}{\emph{Movie reviews}} \\
BoW & 0.79 & 0.80 & 0.88 & 1000\\
BERT & 0.82 & 0.83 & 0.91 & 500\\

 \multicolumn{3}{c}{\emph{Essay}} \\
BoW & 0.97 & 0.80 & 0.99 & 1\\
BERT & 0.98 & 0.87 & 0.99 & 10\\

 \multicolumn{3}{c}{\emph{Hate speech}} \\
BoW      & 0.87 & 0.40 & 0.81 & 10 \\
BERT &  0.89 & 0.63 & 0.88 & 10\\
\multicolumn{3}{c}{\emph{Tweet sentiment}} \\
BoW         & 0.70 & 0.70 & 0.75 & 500\\
BERT        & 0.75 & 0.76 & 0.84 & 1000\\
\hline
\end{tabular}
\caption{The model performance under different datasets.}
\label{table:model-performance}
\end{table}

\subsection{Comparison between removal and relabeling on clean training set}
\label{no-noise-remove-relabel}
When there is no noise in the training set, we run Removal Alg1, Removal Alg2, and Algorithm \ref{alg:alg1} to compare the average returned training set size in Table \ref{table:comapre-remove-label}. It shows that considering training points to relabel can result in smaller training sets than removing them.

\begin{table}[!ht]
\centering
\begin{tabular}{lccc}
\hline
& \textbf{Loan} & \textbf{Reviews} & \textbf{Speech} \\
\hline
Removal Alg1 & 965.4 & 712.8 & 768.6 \\
Removal Alg2 & 440.4 & 636.8 & 411.6 \\
\textbf{Relabeling (ours)} & {\bf 67.0} & {\bf 138.5} & {\bf 49.3} \\
\hline
\end{tabular}
\caption{The comparison of average on $k = |\mathcal{S}_t|$ values over a random subset of test points $x_t$, result by removal (Algorithm 1 and Algorithm 2 \cite{yang2023many}) and relabel. Relabel always finds a smaller $\mathcal{S}_t$ compared with removal.}
\label{table:comapre-remove-label}
\end{table}

\subsection{Running time of Algorithm \ref{alg:alg1}.} 
\label{app:running-time}
We recorded the average running time of Algorithm \ref{alg:alg1} to find $\mathcal{S}_t$ for test points in different datasets in Table \ref{tab:run-time} on Apple M1 Pro CPUs. For one test point, it just takes milliseconds to go through the whole training set (the training set sizes are provided in \ref{app:data-model-details}) to find $\mathcal{S}_t$. 

\begin{table}[h]

\centering
\begin{tabular}{ccc}
\hline
\textbf{Dataset} & \textbf{BoW (ms)} & \textbf{BERT (ms) } \\
\hline
Movie Reviews & 19.04 & 140.51 \\
Essays & 160.01 & 265.09 \\
Hate speech & 103.70 & 299.46 \\

Tweet & 58.42 & 260.75 \\

Loan & 63.97 & / \\
\hline
\end{tabular}
\caption{Average running time (in milliseconds) of Algorithm \ref{alg:alg1} to find $\mathcal{S}_t$ for a test point in different datasets.}
\label{tab:run-time}
\end{table}

\subsection{Full Plots}
\label{sec-full-plots}
We present the distribution of $\mathcal{S}_t$ across various datasets in Tables \ref{fig:hist-app1} and \ref{fig:hist-app2}. Additionally, the correlation between predicted probability and the size of $\mathcal{S}_t$, denoted by $|\mathcal{S}_t|$, for different datasets is showcased in Tables \ref{fig:kp_appendix1} and \ref{fig:kp_appendix2}.

\clearpage
\begin{figure}[thb!]
\centering

\begin{subfigure}{.91\linewidth}
    \centering
    \includegraphics[width=\textwidth]{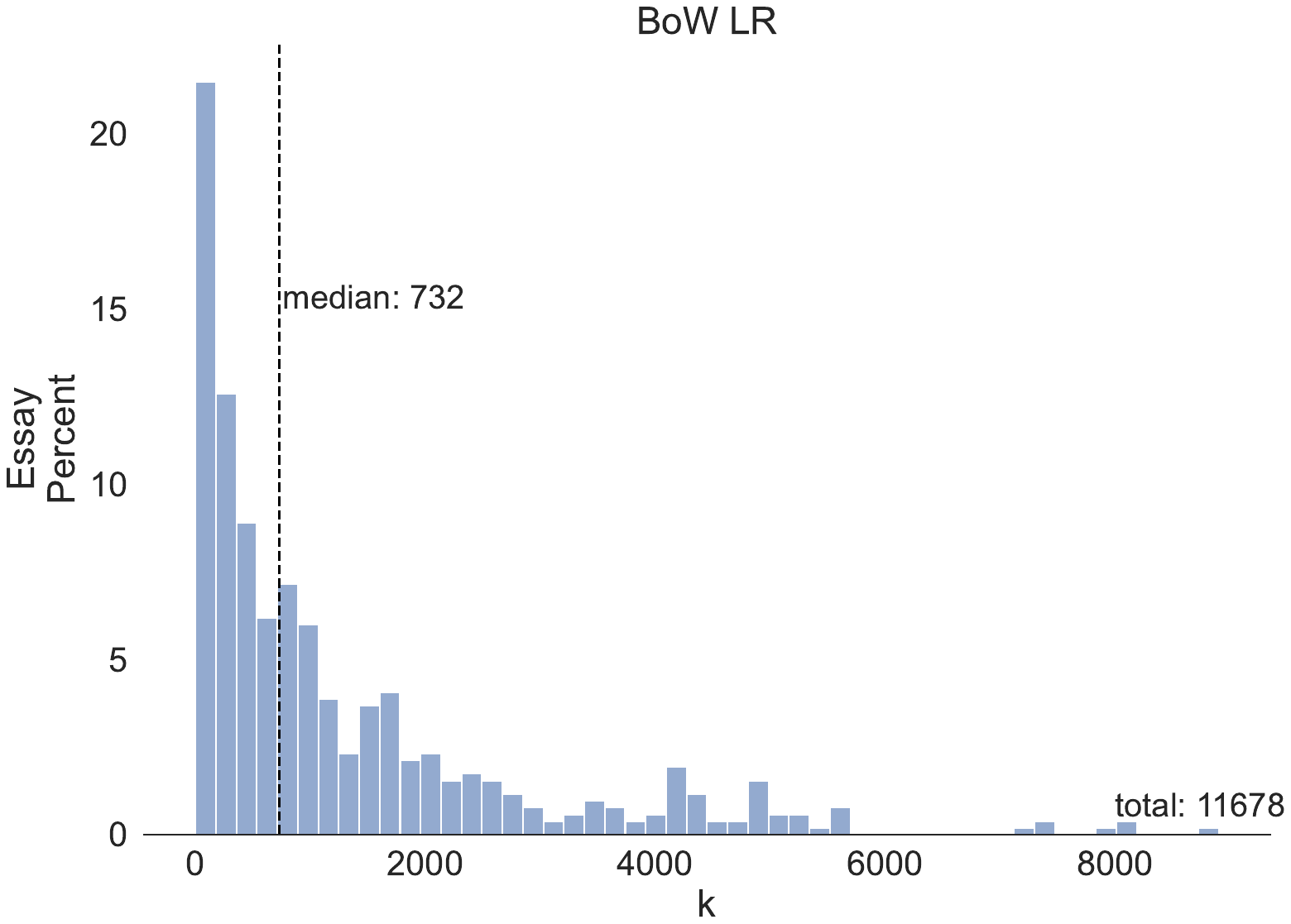}
\end{subfigure}

\begin{subfigure}{.91\linewidth}
    \centering
    \includegraphics[width=\textwidth]{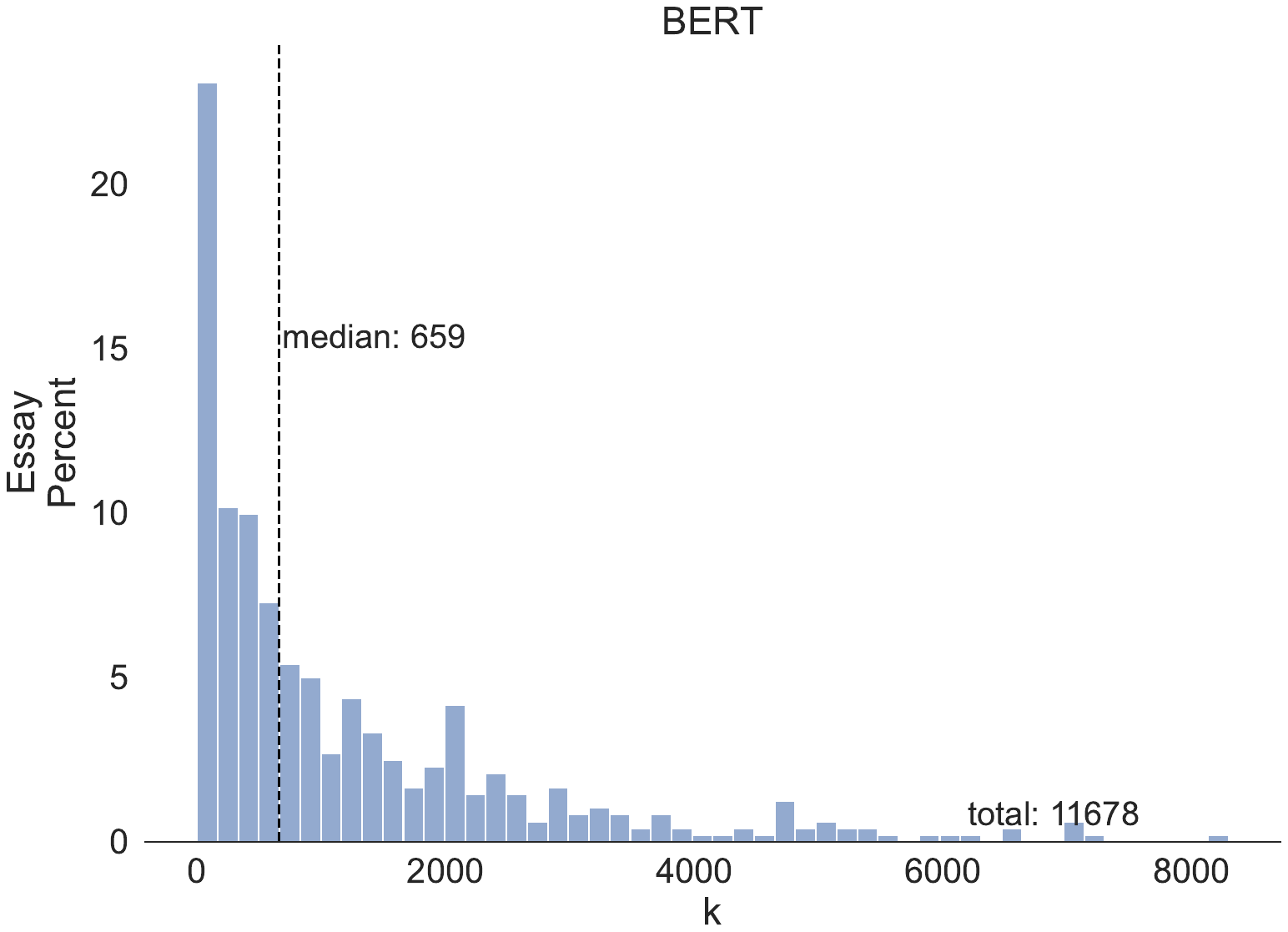}
\end{subfigure}

\begin{subfigure}{.91\linewidth}
    \centering
    \includegraphics[width=\textwidth]{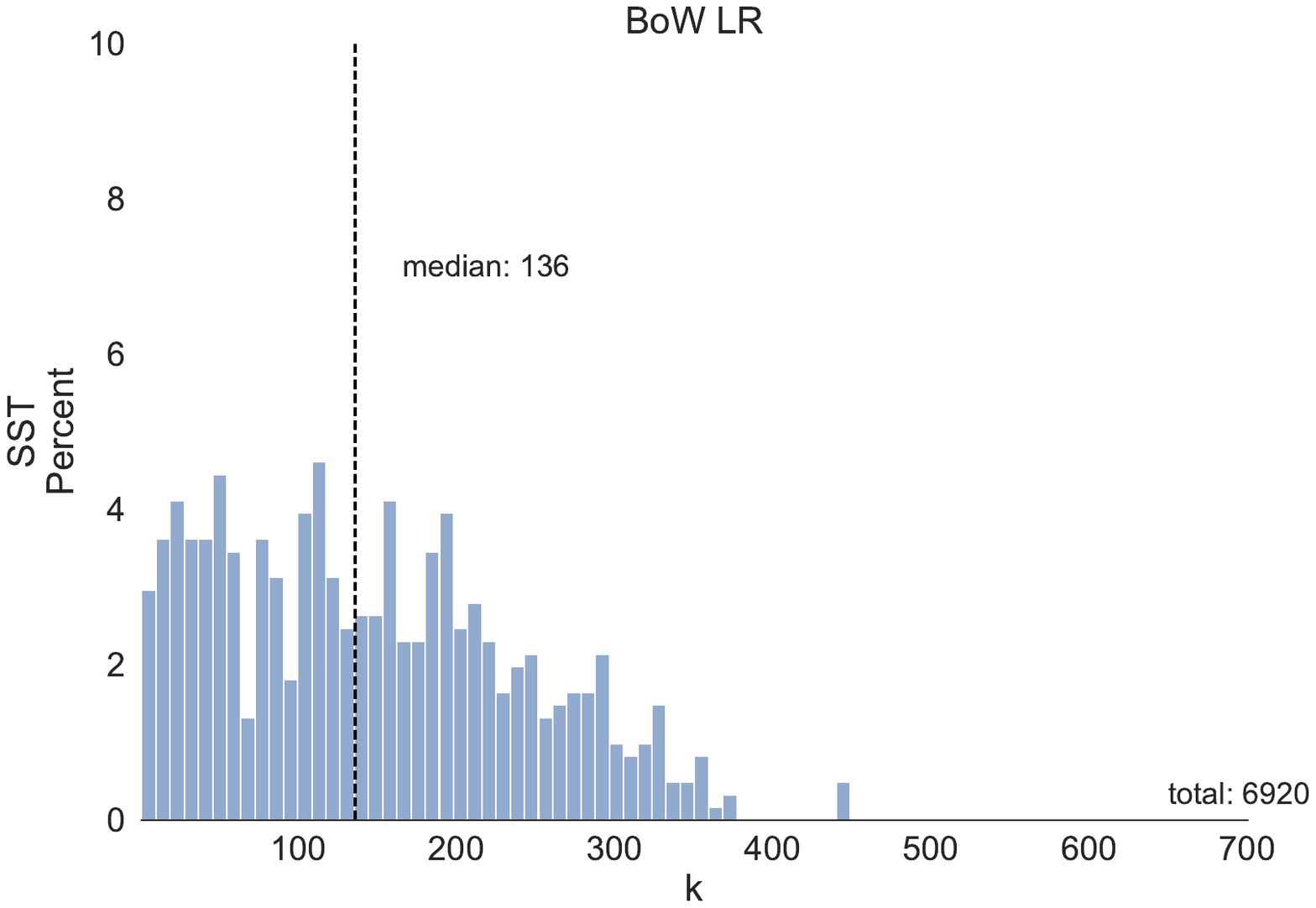}
\end{subfigure}

\begin{subfigure}{.91\linewidth}
    \centering
    \includegraphics[width=\textwidth]{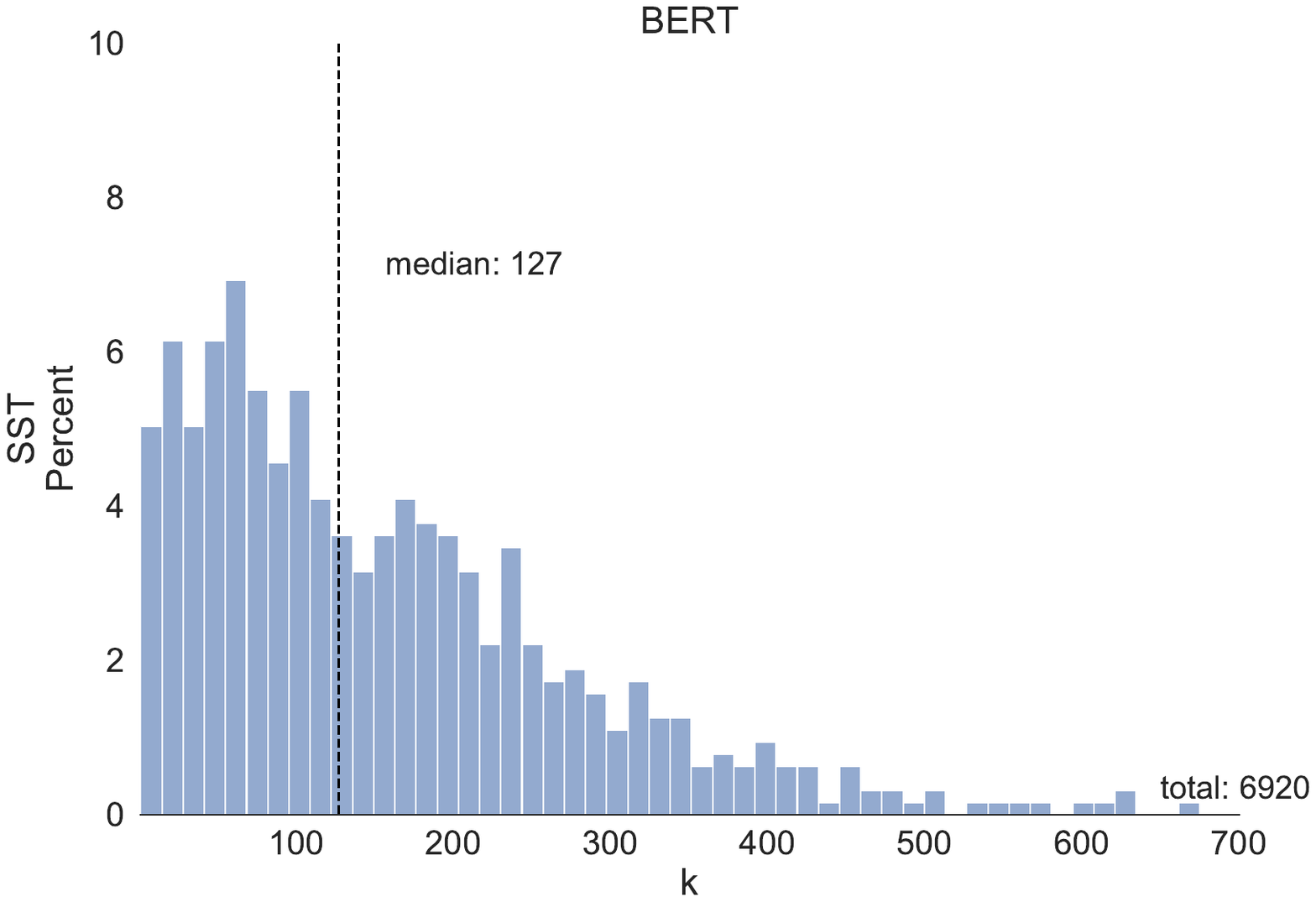}
\end{subfigure}

    \caption{The histogram shows the distribution of $k=|\mathcal{S}_t|$, i.e. the number of points that need to be relabeled from the training data to change the prediction  $\hat{y}_t$ of a specific test example $x_t$.}
    \label{fig:hist-app1}
\end{figure}

\begin{figure}[t!]
\centering

\begin{subfigure}{.86\linewidth}
    \centering
    \includegraphics[width=\textwidth]{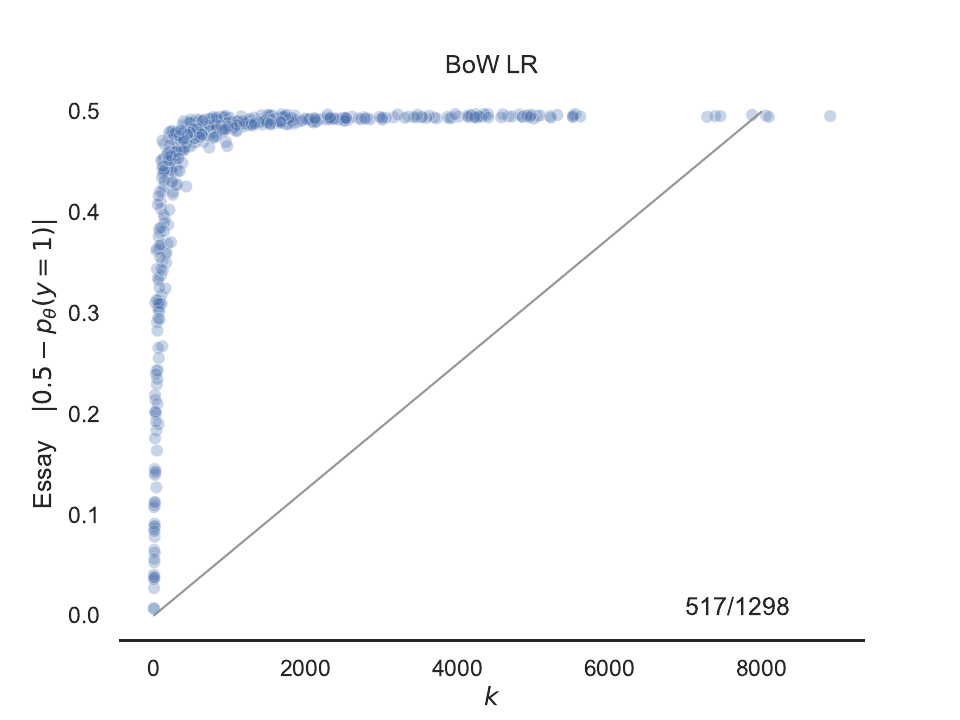}
\end{subfigure}

\begin{subfigure}{.86\linewidth}
    \centering
    \includegraphics[width=\textwidth]{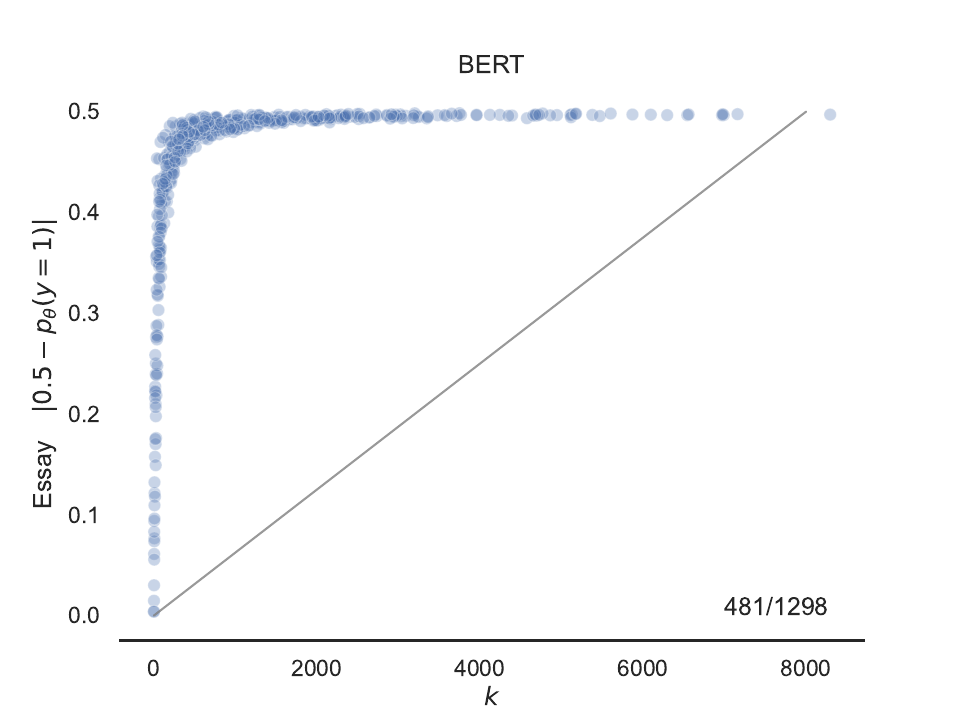}
\end{subfigure}

\begin{subfigure}{.86\linewidth}
    \centering
    \includegraphics[width=\textwidth]{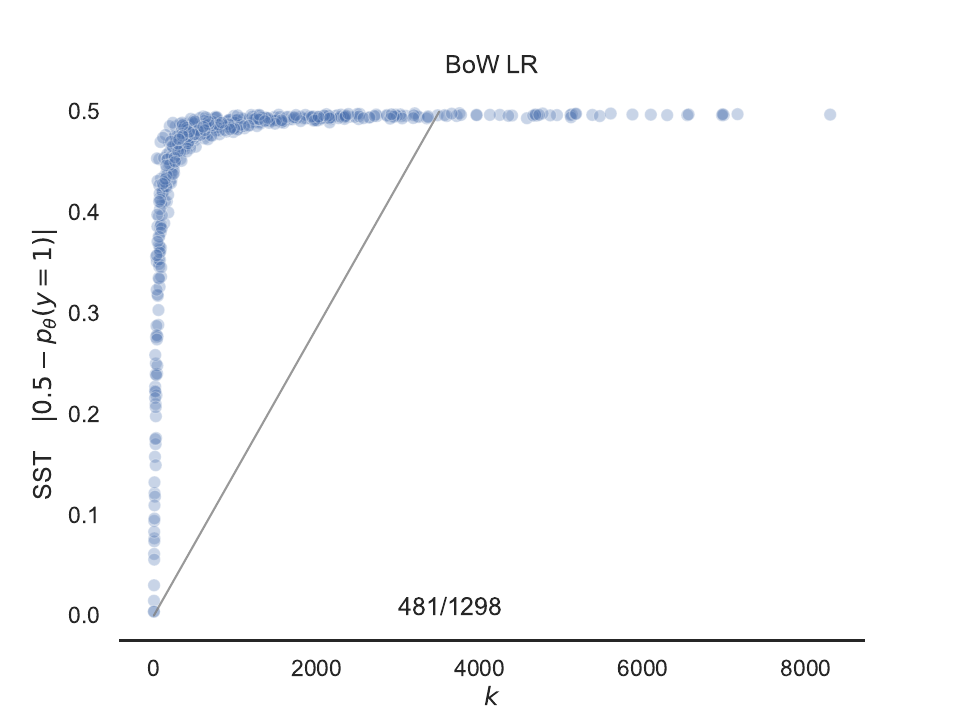}
\end{subfigure}

\begin{subfigure}{.86\linewidth}
    \centering
    \includegraphics[width=\textwidth]{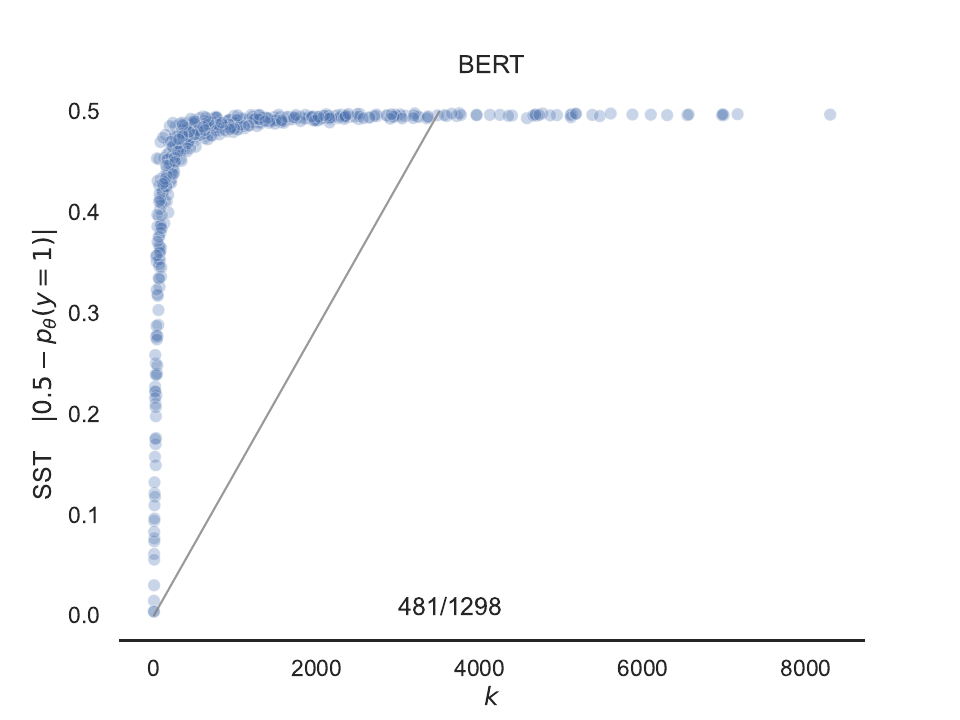}
\end{subfigure}

    \caption{The plot displays the correlation between the predicted probabilities of certain test examples and  $k=|\mathcal{S}_t|$ .There are some test examples where the model is reasonably or highly certain about its prediction, yet by removing a limited number of data points from the training set, the prediction can be altered.}
    \label{fig:kp_appendix1}
\end{figure}

\begin{figure}
\centering
\begin{subfigure}{.91\linewidth}
    \centering
    \includegraphics[width=\textwidth]{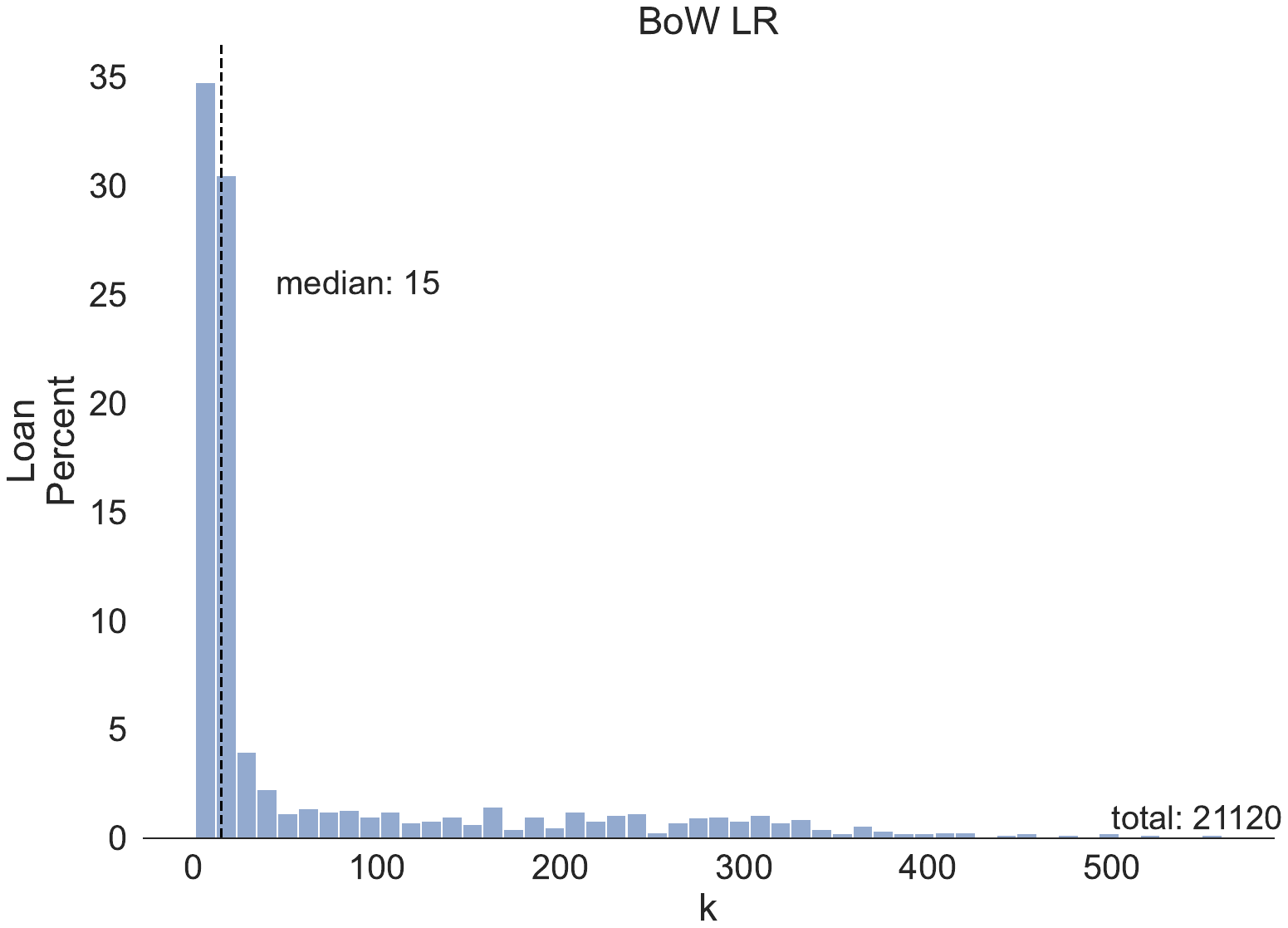}
\end{subfigure}
\begin{subfigure}{.91\linewidth}
    \centering
    \includegraphics[width=\textwidth]{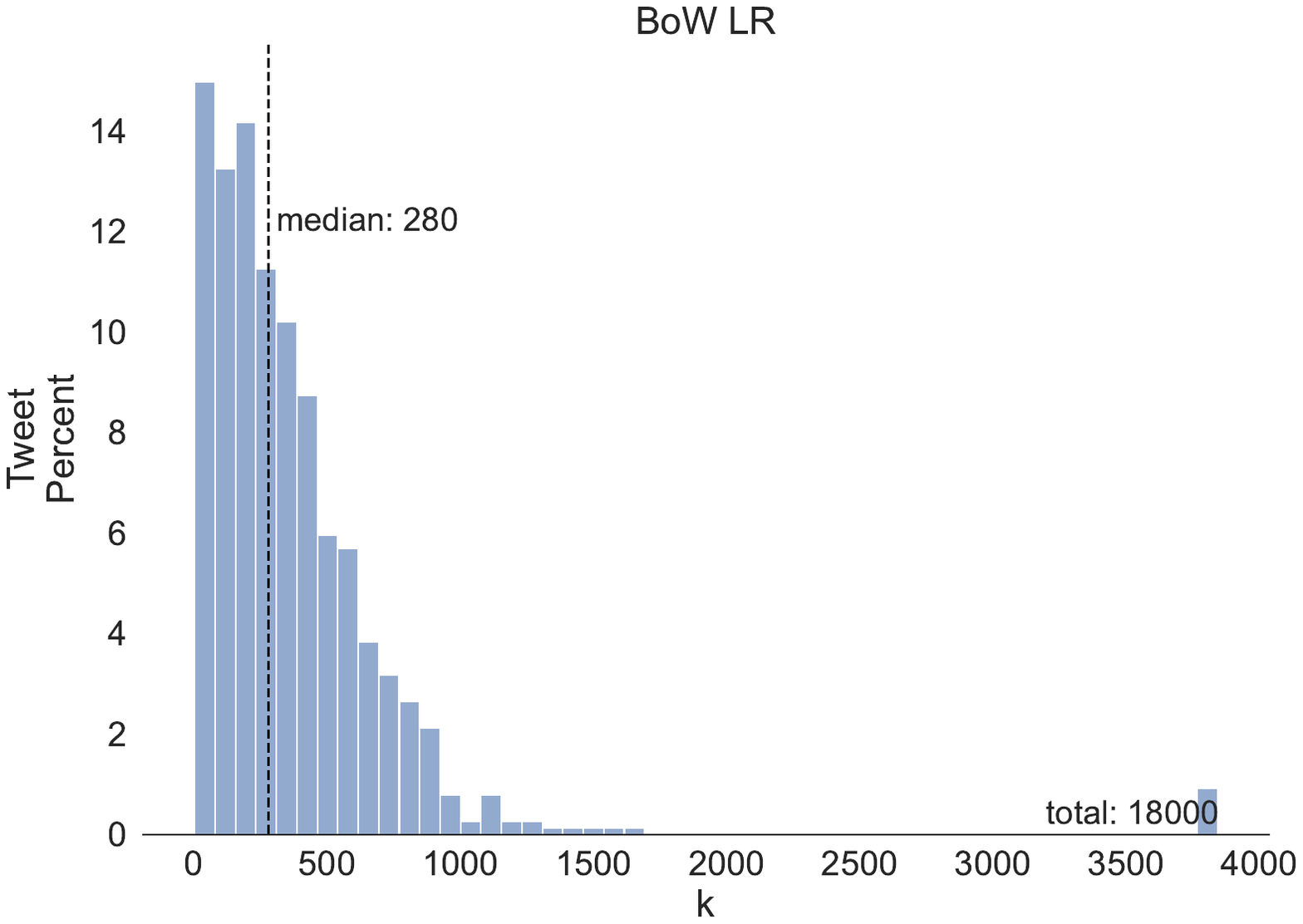}
\end{subfigure}
\begin{subfigure}{.91\linewidth}
    \centering
    \includegraphics[width=\textwidth]{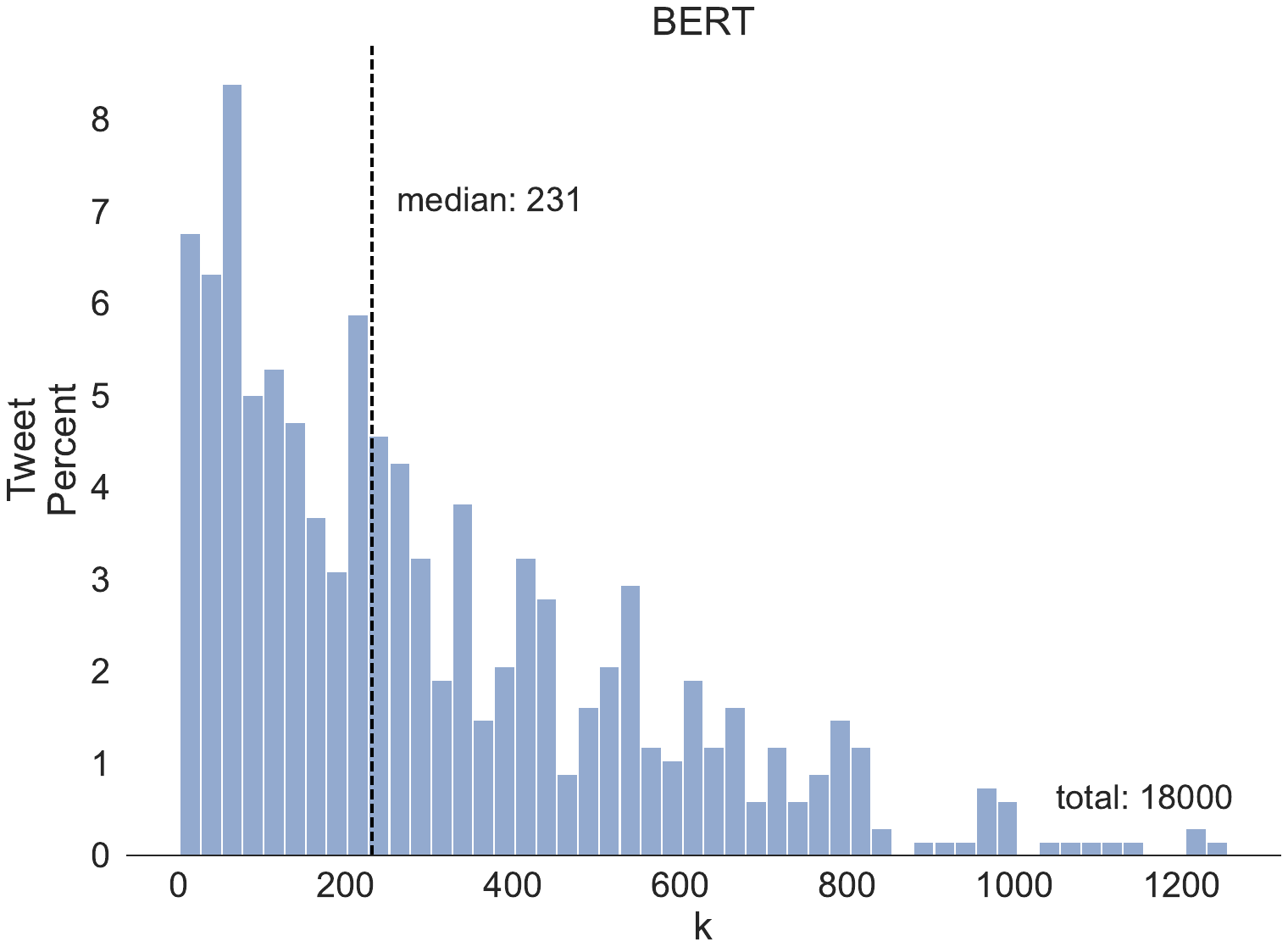}
\end{subfigure}

    \caption{The histogram shows the distribution of $k=|\mathcal{S}_t|$, i.e. the number of points that need to be relabeled from the training data to change the prediction  $\hat{y}_t$ of a specific test example $x_t$.}
    \label{fig:hist-app2}
\end{figure}

\makeatletter
\setlength{\@fptop}{0pt}
\makeatother

\begin{figure}[t!]
\centering

\begin{subfigure}{.86\linewidth}
    \centering
    \includegraphics[width=\textwidth]{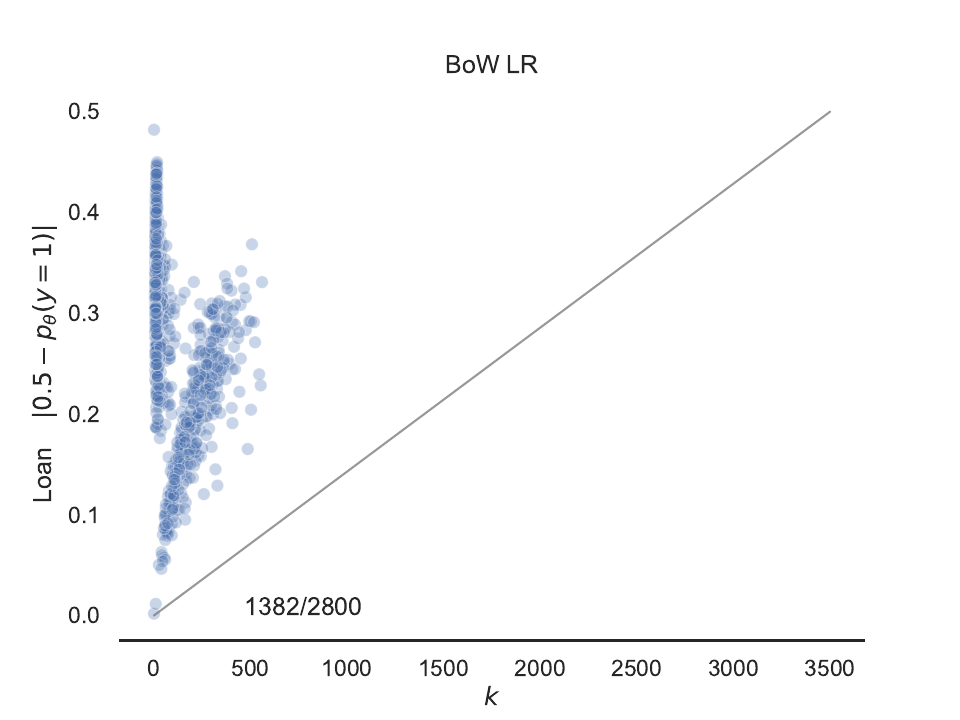}
\end{subfigure}
\begin{subfigure}{.86\linewidth}
    \centering
    \includegraphics[width=\textwidth]{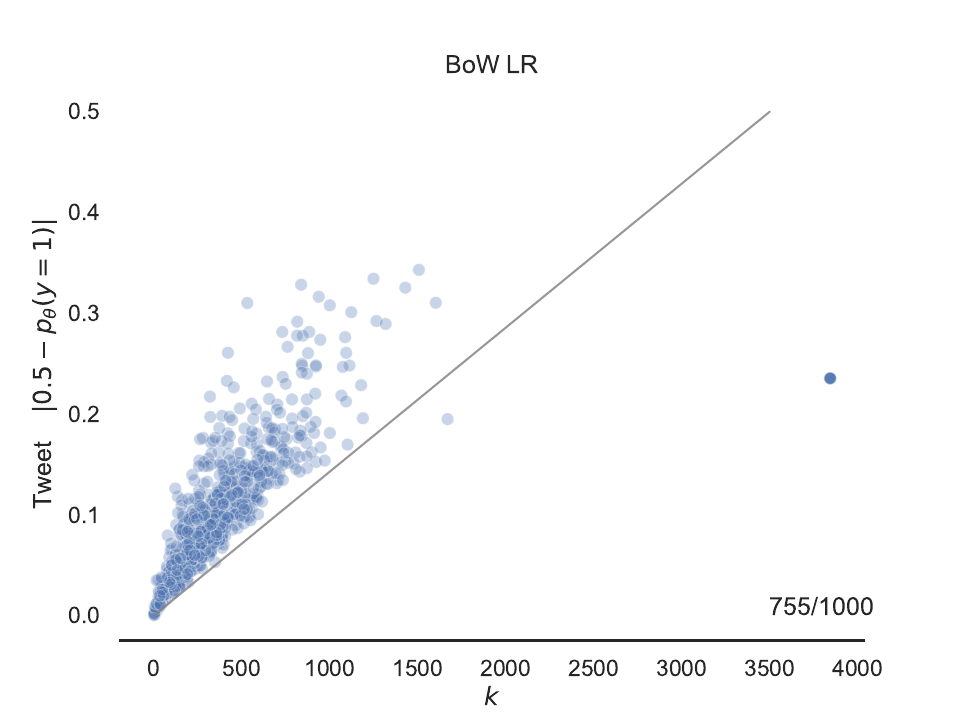}
\end{subfigure}


\begin{subfigure}{.86\linewidth}
    \centering
    \includegraphics[width=\textwidth]{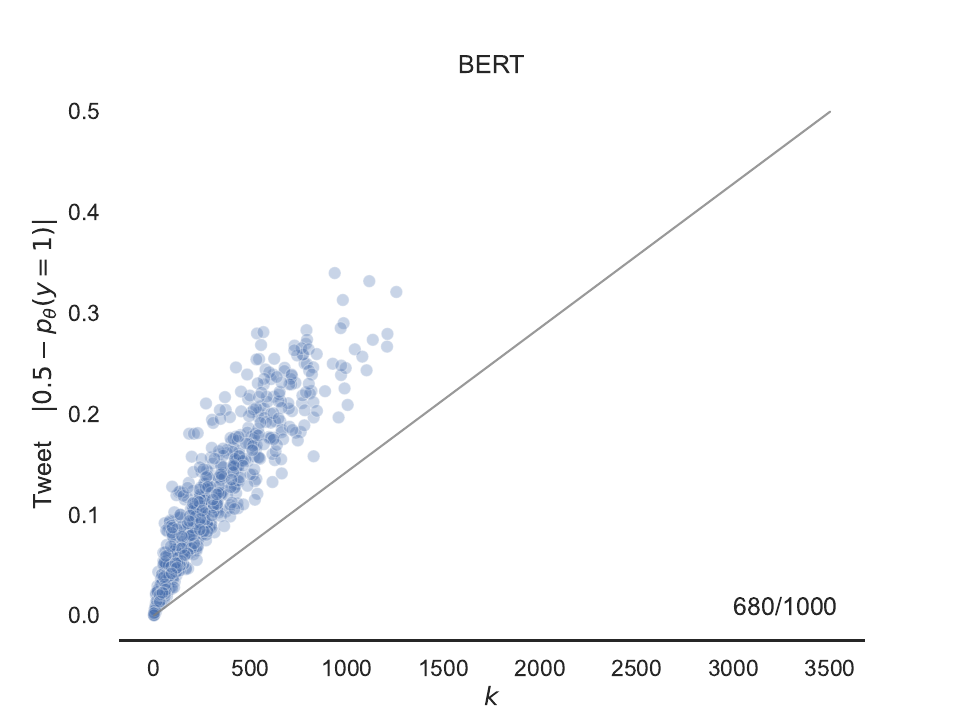}
\end{subfigure}


    \caption{The plot displays the correlation between the predicted probabilities of certain test examples and  $k=|\mathcal{S}_t|$ . Thethere are some test examples where the model is reasonably or highly certain about its prediction, yet by removing a limited number of data points from the training set, the prediction can be altered.}
    \label{fig:kp_appendix2}
\end{figure}

\end{document}